# Convolutional Neural Network Segmentation for Satellite Imagery Data to Identify Landforms Using U-Net Architecture

Mitul Goswami[1], Sainath Dey[1], Aniruddha Mukherjee[1], *Suneeta Mohanty[1], and Prasant Kumar Pattnaik[1]

[1]School of Computer Engineering, Kalinga Institute of Industrial Technology, Bhubaneswar, Odisha, 751024, India

mitulgoswami1212@gmail.com, sainathvia789123@gmail.com, mukh.aniruddha@gmail.com, *suneetamohanty@gmail.com, patnaikprasant@gmail.com

**Abstract.** This study demonstrates a novel use of the U-Net architecture in the field of semantic segmentation to detect landforms using preprocessed satellite imagery. The study applies the U-Net model for effective feature extraction by using Convolutional Neural Network (CNN) segmentation techniques. Dropout is strategically used for regularization to improve the model's perseverance, and the Adam optimizer is used for effective training. The study thoroughly assesses the performance of the U-Net architecture utilizing a large sample of preprocessed satellite topographical images. The model excels in semantic segmentation tasks, displaying high-resolution outputs, quick feature extraction, and flexibility to a wide range of applications. The findings highlight the U-Net architecture's substantial contribution to the advancement of machine learning and image processing technologies. The U-Net approach, which emphasizes pixel-wise categorization and comprehensive segmentation map production, is helpful in practical applications such as autonomous driving, disaster management, and land use planning. This study not only investigates the complexities of U-Net architecture for semantic segmentation, but also highlights its real-world applications in image classification, analysis, and landform identification. The study demonstrates the U-Net model's key significance in influencing the environment of modern technology.

# 1 Introduction

Deep learning has emerged as a transformational force in artificial intelligence and computer vision, revolutionizing picture categorization and analysis. Inspired by the architecture and operation of the human brain, this ground-breaking subset of machine learning is characterized by multilayer neural networks that enable hierarchical feature extraction and abstraction. [1]. Because of the inherent complexity and variety of visual data, this approach has historically been difficult [2]. Deep learning, on the other hand, has enabled the development of CNNs that can learn and extract key features from pictures automatically, significantly boosting classification performance. These models can recognize complex patterns, textures, and forms, making them invaluable for applications like item detection, face recognition, and medical picture analysis [3].

Landform recognition and mapping are crucial in various fields, from autonomous navigation to environmental monitoring and disaster response. It involves understanding and depicting the physical characteristics of the Earth's surface, which can vary greatly in terms of topography, vegetation, and other features [4]. Modern techniques for landform recognition and mapping often rely on remote sensing technologies like satellite imagery, LiDAR (Light Detection and Ranging), and aerial photography [4] [5]. Additionally, advanced models like the U-Net architecture, as mentioned earlier, enable precise localization of landform classes and parameters like roughness and slipperiness. Because of its capacity to acquire complete images of the Earth's surface, imagery from satellites is crucial in landform detection [6].

Satellite imagery's precision and regular updates enable professionals to make educated judgments, leading to advances in geography, environmental research, and sustainable land management practices [7]. This study describes an innovative U-Net-based CNN architecture for precise semantic segmentation and classification of preprocessed satellite images. Its novel contribution is the development of terrain recognition, which reliably identifies varied landforms such as forests, deserts, mountains, and bodies of water, with implications for defense requirements, environmental monitoring, and disaster response.

# 2 Related Works

In recent years, advancements in the application of CNNs, particularly using the U-Net architecture, have significantly enhanced the capabilities of satellite imagery data segmentation for identifying various landforms and features. Bagaev et al. [8]

demonstrated the effectiveness of U-Net in segmenting Earth's surface images, focusing on identifying distinct landforms. Unlike their broader approach, our research employs U-Net specifically for nuanced landform identification, optimizing feature extraction for more precise segmentation. Similarly, Kattenborn et al. [9] achieved high accuracy in vegetation mapping using U-Net, illustrating its precision in ecological segmentation tasks. Our work diverges here, concentrating on landforms rather than vegetation, and introducing a specialized approach to dropout and optimizer usage to enhance segmentation accuracy. Tiurin et al. [10] extended U-Net's application to the segmentation of various features such as agricultural fields and urban buildings, showcasing its versatility in handling diverse landscapes. In contrast, our paper narrows the focus solely to landforms, applying U-Net in a more targeted fashion. The work by Gonzales [11] on cloud detection in satellite imagery using a deep convolutional U-Net architecture with transfer learning further demonstrates the model's adaptability to different segmentation challenges. Our study aligns with this adaptability but redirects it toward the specific challenge of landform identification.

Liu et al. [12] introduced the D-RESUNET, a variant of U-NET combined with residual learning and dilated convolutions, achieving superior performance in road area extraction, indicating its effectiveness in urban planning applications. Our research, although similarly innovative, diverges in its application, focusing on natural rather than urban landscapes. In a different domain, Chen et al. [13] developed DRINET, an architecture that outperforms U-Net in medical image segmentation, suggesting potential cross-disciplinary applications of these advancements. This multidisciplinary potential is echoed in our work, though we harness it specifically for satellite-based landform segmentation. Grosgeorge et al. [14] combined U-Net with RetinaNet for enhanced aircraft detection in satellite images, indicating the architecture's utility in specific object detection scenarios. Our study, while also exploring the utility of U-Net in specific scenarios, concentrates on the semantic segmentation of landforms rather than object detection. Sedov et al. [15] explored optimizing U-Net with different loss functions for residential area segmentation, contributing to the nuanced understanding of network tuning for specific segmentation tasks. Our paper contributes similarly to the understanding of U-Net's tuning, but in the context of landform identification, focusing on the efficacy of specific dropout and optimizer strategies.

Furthermore, Gonzalez et al. [16] demonstrated U-Net's application in segmenting water bodies, underscoring its effectiveness in diverse natural feature identification. Our research complements this by applying U-Net to a wider range of landforms, showing its versatility in a different domain. Lastly, Ji et al [17] enhanced U-Net's performance for building extraction using the Siamese U-Net architecture, proving its efficacy in large structure identification in remote sensing imagery. Our research stands apart in its focus on natural landscapes, demonstrating U-Net's efficacy in landform segmentation with a specialized approach.

These advancements collectively illustrate the expanding scope and sophistication of U-Net in satellite imagery segmentation, setting a foundation for future innovations in landform identification and beyond. This paper builds on this foundation, focusing on a direct application of U-Net for landform identification in satellite images, emphasizing feature extraction and semantic segmentation with a specialized approach to dropout and optimizer usage.

## 3 U-Net Architechture

The U-Net architecture, distinguished by its characteristic U-shaped structure, is a strong and adaptable deep learning model that was created for image segmentation tasks. However, because of its versatility and remarkable feature extraction capabilities, it is well-suited for image categorization and landform identification applications. U-Net's architecture is composed of two primary components: an encoder and a decoder, connected through skip connections. The encoder, located on the left side of the U, is responsible for downsampling and feature extraction. It consists of a series of convolutional layers with increasing receptive fields. These layers process the input image, extracting essential features at different levels of abstraction [18]. The hierarchical feature extraction is fundamental to the model's ability to understand the context of the image. The skip connections, a defining feature of U-Net, create connections between corresponding layers in the encoder and decoder[19].

In the context of image classification, U-Net can be adapted by modifying the output layer. The encoder, renowned for its efficient feature extraction, serves as a feature extractor. These features are then passed to a fully connected layer, which produces a class prediction for the entire image. U-Net's strength in image classification becomes evident when applied to fine-grained recognition tasks, where it excels at recognizing specific objects or regions within images [20]. The combination of detailed feature extraction and spatial context retention through skip connections results in high classification accuracy.

U-Net is a valuable tool for landform recognition when trained on a dataset of landform images and corresponding labels. In landform recognition, U-Net segments an input image into distinct landform classes, producing a detailed map of landform distribution within the scene. This fine-grained understanding of the environment is essential for making informed decisions in applications like autonomous driving, disaster management, and land use planning. U-Net's architecture offers several advantages in image classification and landform recognition. Furthermore, U-Net's flexibility allows it to be adapted for various tasks, making it a versatile tool for different applications. Its pixel-wise classification and segmentation capabilities frequently result in highly accurate outputs, which are particularly valuable in tasks that demand precision [21].

## 4      Model Implementation

As the input data source for this research, a comprehensive dataset of preprocessed satellite images was used. The fundamental goal of this research is to create a reliable framework that can generate segmented pictures from raw satellite imagery. This segmentation procedure entails categorizing each pixel in the images into several landscape classes. The model's implementation, based on the U-Net architecture, is critical to attaining this aim.

The proposed methodology combines sophisticated algorithms for semantic segmentation and content-based image retrieval (CBIR). The algorithmic representation starts with preprocessing satellite images. Feature extraction utilizes a CNN architecture based on the U-Net model. The CNN includes an encoder-decoder structure, Conv2dBlock for hierarchical feature extraction, and skip connections for spatial context retention. The decoder path involves upsampling using Conv2DTranspose layers. The resulting features are then employed for semantic segmentation. In parallel, for CBIR, binary hashing is applied to convert extracted features into unique binary hash codes. During retrieval, query features undergo the same hashing process for matching. This unified approach enhances both segmentation accuracy and efficient image retrieval, connecting the realms of image analysis and retrieval within a comprehensive algorithmic framework.

The model consists of an encoder-decoder architecture with Conv2dBlock as the building block. The Conv2dBlock involves two convolutional operations, batch normalization, and ReLU activation. The GiveMeUnet function creates the U-Net model with specified filter sizes, dropout rates, and batch normalization. The encoder path progressively reduces spatial dimensions and increases feature channels through convolution and max-pooling operations. The decoder path then upsamples the features using Conv2DTranspose layers and concatenates them with corresponding encoder feature maps using skip connections. These skip connections enable the network to leverage low-level characteristics for fine-grained segmentation. The use of ReLU activation in Conv2dBlock prevents the vanishing gradient problem and encourages sparsity, enhancing the generalization performance. Dropout is employed for regularization, and the final layer uses sigmoid activation for pixel-wise segmentation output (see Fig. 1).

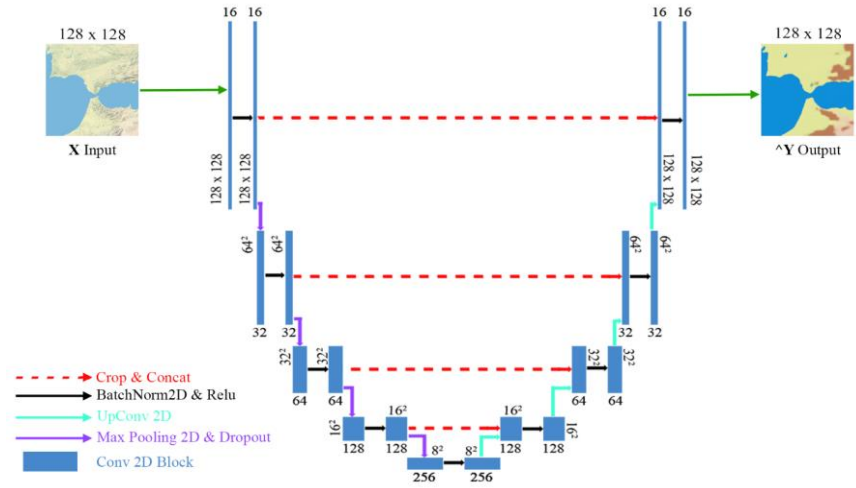

**Fig. 1.** U-Net Model Flow Diagram

Let *U-Net(x)* represent the U-Net architecture's forward pass on input *x*, and *F* be the set of operations applied in each block. The equation can be roughly summarized as:

$$U - Net(x) = F(F(F(F(x)))) -> \ldots -> F(x)) \qquad (1)$$

Equation (1) highlights the recursive and hierarchical nature of the U-Net architecture, where the input undergoes a series of operations (F) at multiple levels, with skip connections connecting the encoder and decoder paths. In the last layer, the sigmoid activation function is employed to generate pixel-wise binary segmentation masks. It compresses the network's output values into the range [0, 1], where values closer to 1 indicate the existence of the target item and values closer to 0 indicate the absence of the target object.

$$S(x) = \frac{1}{(1+e^{-x})} \qquad (2)$$

S(x) in equation (2) is the sigmoid function output for input x. e is the natural logarithm's base, and it is approximately equivalent to 2.71828. When x is a real integer in this equation, the sigmoid function transfers it to the range (0, 1). S(x) approaches 1 as x gets more positive, while S(x) approaches 0 as x grows more negative. The model produces a segmented picture in which each pixel is assigned a value between 0 and 1, indicating the likelihood that it corresponds to the target item. The U-Net model employs dropout with a dropout rate of 0.07. Dropout is a regularization method used to prevent overfitting in neural networks.

For training the model, the Adam optimizer is utilized. It's a well-known optimization procedure that combines the advantages of two existing optimization techniques: Adagrad and RMSprop. Adam is well-known for its efficiency and capacity to adjust the learning pace while in training. To compute an adaptive learning rate for each parameter, it keeps moving averages of previous gradients and past squared gradients. This enables the model to converge more quickly and perform better across a wide range of tasks. The complete update equation for the Adam optimizer, in a more concise form, can be expressed as a single equation:

$$\theta_t = (\theta_{t-1} - \alpha) * \frac{m_t}{\sqrt{v_t} + \epsilon} \qquad (3)$$

In equation (3), $\theta_t$ represents the updated parameter vector at time step t. The optimizer is iteratively updating the model's parameters during training to minimize the loss. $\theta_{t-1}$ represents the previous parameter vector, representing the values of the model's parameters at the previous time step (t-1). $\alpha$ is the learning rate, a hyperparameter that controls the step size during the parameter updates. It determines how much the model's parameters are adjusted in each iteration. A smaller learning rate leads to more stable but slower convergence, while a larger learning rate can lead to faster convergence but might overshoot the optimal solution. $m_t$ is calculated as a moving average of the gradients of the loss concerning the parameters. The β1 hyperparameter determines the exponential decay rate for first-moment estimations. This word describes a momentum-like effect in the optimization process. The second-moment estimate, $v_t$, is calculated using a moving average of the squared gradients. It aids in adjusting the learning rates for each parameter independently. The β2 hyperparameter determines the exponential decay rate for second-moment estimations. ε is a tiny constant added to the denominator to avoid division by zero and provide numerical stability.. The Adam optimizer combines these terms to update the model's parameters efficiently [22]. The binary cross-entropy loss is used in conjunction with the U-Net segmentation model to penalize the model for erroneous predictions regarding the presence or absence of a target item in each pixel of the image. The model is trained to minimize this loss function, which in turn improves its ability to accurately segment the target object. The use of binary cross-entropy as a loss function is particularly appropriate for U-Net models because they are designed to produce pixel-level segmentation masks. The binary cross-entropy loss for the pixel is calculated as:

$$L = -y * log(p) - (1 - y) * log(1 - p) \qquad (4)$$

In equation (4), $y$ is the actual class label (0 or 1), and p is the predicted probability. This formula effectively penalizes the model for making inaccurate predictions. If the model predicts a high probability (p close to 1) for a pixel that belongs to the object (y = 1), the loss will be low. Conversely, if the model predicts a low probability (p close to 0) for a pixel that belongs to the object, the loss will be high. The same applies to pixels that do not belong to the object (y = 0). By minimizing the binary cross-entropy loss during training, the U-Net model learns to produce more accurate segmentation masks, effectively distinguishing between target objects and the background. The Sørensen-Dice index, or Dice coefficient, is a statistic used to assess how well image segmentation algorithms work. It is a measurement of how closely the expected and ground truth segmentations overlap. The Dice coefficient is calculated as follows:

$$Dice\ Coefficient = \frac{2*TP}{2*TP+FN+FP} \qquad (5)$$

The anticipated and ground truth segmentations perfectly match when the Dice coefficient is 1, which has a range of 0 to 1. When the Dice coefficient is zero, there is no overlap; when it is 0.5, there is some overlap, but not perfect overlap. In equation (5) *TP* denotes True Positive, *FN* denotes False Negative and *FP* denotes False Positive. In the realm of image segmentation, the Dice coefficient stands out as a superior metric compared to Intersection over Union (IoU) or accuracy [23]. Its sensitivity to false positives, resilience to class imbalance, emphasis on boundary detection, a direct link to precision and recall, and ease of interpretation set it apart. The Dice coefficient's strength lies in its ability to penalize false positives more severely than IoU, a crucial aspect in medical imaging where false positives can lead to misdiagnosis or mistreatment.

## 5  Experimentations and Results

To evaluate the model's performance, a comprehensive dataset of 5,000 preprocessed satellite landform images is employed. Each image comprises a randomized 512x512 pixel segment of the Earth's surface and consists of three components: a Landform map, a Height map, and a Segmentation map. The input images represent landform maps of the Earth that have been segmented to identify distinct landforms such as deserts, forests, mountains, and other geological features. The detailed pictorial comparison of the output of the semantic segmentation has been represented (see Fig. 2).

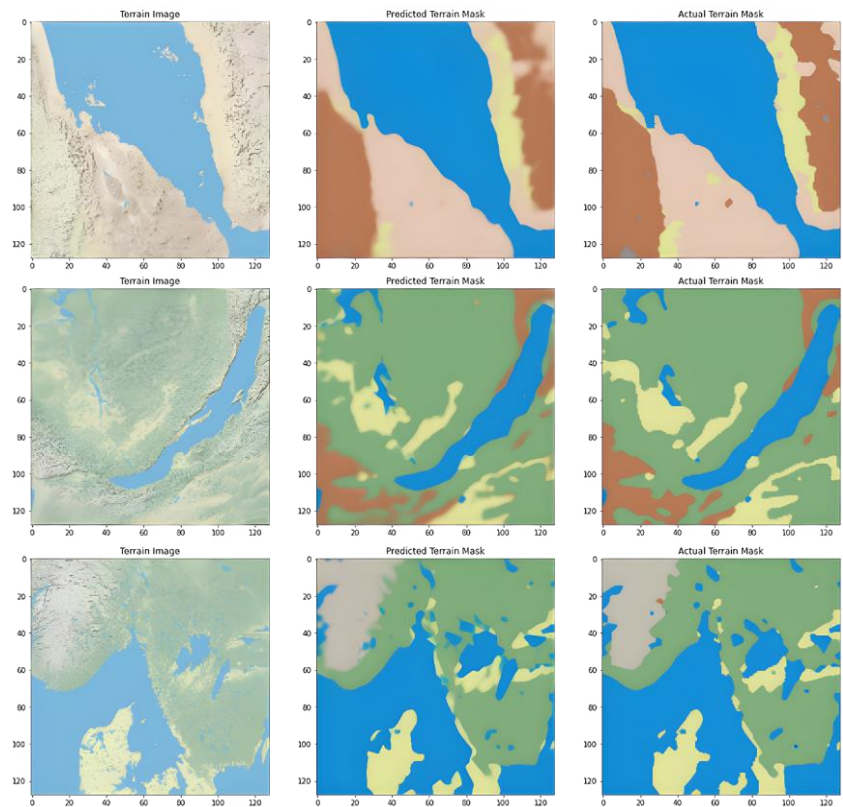

**Fig. 2.** Comparing Raw Landform Data, the Actual Landform Mask, and the Predicted Landform Mask

The acquired findings provide convincing proof of the U-Net model's competence in image segmentation. It effectively parses the subtle features in the raw landform images and produces a segmented result with an impressive level of accuracy. This potential is seen when the predicted images are compared to the real landform mask, revealing just slight, inconspicuous differences that have no substantial influence on the overall quality of the segmentation. The model's pixel-to-pixel segmentation technique appears to be extremely efficient, retaining the integrity of the landforms portrayed in the input image. The segmented image that results utilizes a well-chosen color palette to visually differentiate the many types of landforms. For example, the lush green color palette is well used to convey the vastness of forests and grassy fields. The yellow color, on the other hand, reflects the sandy landform. Furthermore, the peach color gives the signal for mountains and hilly regions. This thorough segmentation method not only improves image quality but also assures the portrayal of Earth's numerous landform characteristics. The above results are obtained upon training the model on 5000 images of each of the segmented maps and landform maps. However, in real-life applications, the datasets contain images in hundreds of thousands which maximizes the potential output, almost indistinguishable from the ground truth. For the evaluation process, we employ a combination of visual inspection along with the Dice coefficient metric and Model Accuracy of the model to gauge the pixel-level accuracy between the model's output and the ground truth [24]. This dual approach allows for comprehensively assessing the model's performance, ensuring a thorough and effective evaluation of its segmentation accuracy. Visual inspection provides a qualitative assessment, while the Dice coefficient and Model Accuracy offers a quantitative measure, together offering a holistic view of the model's ability to accurately capture and delineate objects or regions within the images.

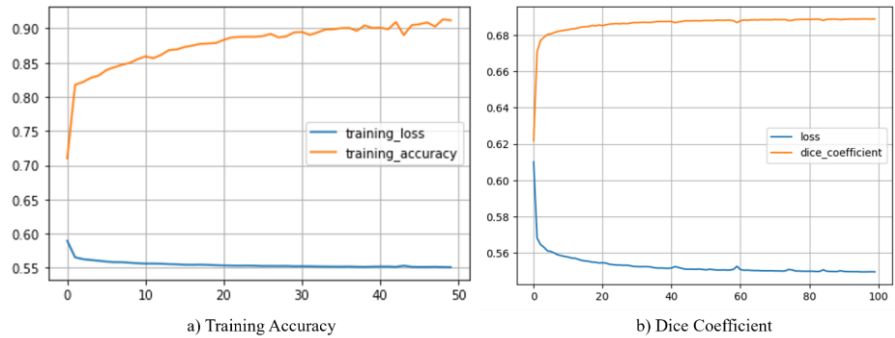

**Fig. 3.** Performance Metrics of the U-Net Model

The model achieves a commendable Dice Coefficient of 69.62% and a Model Accuracy of 90.53% (see Fig. 3), signifying a robust agreement between the predicted and ground truth segmentations for the targeted class. This metric, evaluating pixel-wise overlap, indicates that the model accurately identifies around 69% of the pixels assigned to the class in the ground truth. In segmentation tasks, a Dice value of 69.62% is widely considered excellent, highlighting the model's strong segmentation capabilities. The notable Model Accuracy of approximately 90.53% underscores the model's proficiency in learning and reproducing patterns from the training dataset. This metric represents the percentage of correctly classified instances during the training phase, indicating a high level of convergence and model fitting. The Model Accuracy suggests that the model effectively captures the underlying features and relationships within the training data, demonstrating its capacity to generalize well to seen examples. The performance of the proposed CNN-based U-Net segmentation has been meticulously compared with other state-of-the-art models and architectures. This comprehensive evaluation provides a thorough understanding of the proposed model's effectiveness in relation to existing cutting-edge approaches in the field. (see Table. 1).

**Table 1.** Evaluation of the proposed model's performance metrics

| Method | Dice Coefficient (%) |
|---|---|
| CNN based U-Net (Ours) | 69.62 |
| D – RESUNET [12] | 62.93 |
| DRINet [13] | 83.42 – 96.57 |
| SiU-Net [17] | 59.50 – 61.10 |

The CNN-based U-Net segmentation outperforms comparative models, achieving a Dice Coefficient of 69.62%. Notably, DRINet records a wide performance range from 83.42% to 96.57%, demonstrating its variability. D-RESUNET achieves 62.93%, while SiU-Net ranges between 59.50% and 61.10%. The superior performance of the proposed model underscores its efficacy in semantic segmentation, offering a promising solution for accurate and robust image analysis tasks compared to state-of-the-art alternatives.

# 6 Conclusion

In conclusion, the study gives a thorough examination of deep learning's transformational influence, notably the U-Net architecture, in the realms of picture classification, analysis, and landform identification. Deep learning, which is inspired by the structure of the human brain, has considerably improved the accuracy and precision of tasks like as picture categorization. Because of its distinctive U-shaped structure, the U-Net architecture emerges as a versatile and effective model for image segmentation, classification, and landform recognition addressing various tasks within a unified framework. The encoder-decoder structure, together with skip connections, enables the exact segmentation of complicated pictures by facilitating hierarchical feature extraction and spatial context retention enabling precise segmentation of complex images while retaining spacial context. The use of the Adam optimizer contributes to efficient training by dynamically adapting learning rates, allowing the model to converge quickly across a range of tasks. The choice of binary cross-entropy as the loss function aligns with U-Net's pixel-wise segmentation nature, guiding the model to produce accurate segmentation masks. On the other hand, The use of the Dice coefficient as a class-specific metric demonstrates a substantial agreement (69%) between predicted and ground truth masks, providing a reliable measure of segmentation performance. The U-Net architecture's effectiveness in landform recognition lies in its ability to produce high-resolution outputs, efficient feature extraction, and adaptability to various tasks.

The implementation details of the U-Net model highlight its encoder-decoder structure, the role of convolutional blocks in feature extraction, and the use of dropout for regularization. However, The U-Net architecture, with its intricate encoder-decoder structure, may pose computational challenges, particularly in resource-constrained environments. The effectiveness of the proposed approach heavily relies on the availability and quality of labeled training data, potentially limiting its applicability in scenarios with sparse or insufficient data and the performance of the model is sensitive to hyperparameters, such as learning rates and dropout rates, requiring careful tuning for optimal results.

Moreover, deploying the proposed model for real-time applications presents computational, latency, and memory-related challenges. Adapting to different contexts, protecting data privacy, and integrating smoothly with current systems are all important issues. Continuous model updates, interpretability, and addressing edge circumstances all add complexity. Striking a balance between model complexity and real-world performance, resolving latency difficulties, and implementing effective security measures are all critical for successful real-time deployment, necessitating a comprehensive and adaptable strategy to solve these obstacles. Future research scope includes optimizing the proposed U-Net model for edge devices to enhance real-time Additionally, investigating transfer learning strategies and extending the model's capabilities for multi-modal satellite data could broaden its applicability in diverse remote sensing applications.

**References**

1. Janiesch, C., Zschech, P., Heinrich, K.: Machine learning and deep learning. Electron Markets 31(6), 685–695 (2021). https://doi.org/10.1007/s12525-021-00475-2
2. Wang, P., Fan, E., Wang, P.: Comparative analysis of image classification algorithms based on traditional machine learning and deep learning. Pattern Recognition Letters 141, 61–67 (2021). https://doi.org/10.1016/j.patrec.2020.07.042
3. Vijayan, A., Kareem, S., Kizhakkethottam, J.J.: Face Recognition Across Gender Transformation Using SVM Classifier. Procedia Technology 24, 1366–1373 (2016). https://doi.org/10.1016/j.protcy.2016.05.150
4. Kozlowski, P., Walas, K.: Deep neural networks for landform recognition task. 2018 Baltic URSI Symposium (URSI), Poznan, Poland, pp. 283–286 (2018). https://doi.org/10.23919/URSI.2018.8406736
5. Kutila, M., Pyykönen, P., Ritter, W., Sawade, O., Schäufele, B.: Automotive LIDAR sensor development scenarios for harsh weather conditions. 2016 IEEE 19th International Conference on Intelligent Transportation Systems (ITSC), Rio de Janeiro, Brazil, pp. 265–270 (2016). https://doi.org/10.1109/ITSC.2016.7795565
6. Sklansky, J.: Image Segmentation and Feature Extraction. IEEE Transactions on Systems, Man, and Cybernetics 8(4), 237–247 (1978). https://doi.org/10.1109/TSMC.1978.4309944
7. Goswami, M., Panda, N., Mohanty, S., Pattnaik, P.K.: Machine Learning Techniques and Routing Protocols in 5G and 6G Mobile Network Communication System - An Overview. 2023 7th International Conference on Trends in Electronics and Informatics (ICOEI), Tirunelveli, India, pp. 1094–1101 (2023). https://doi.org/10.1109/ICOEI56765.2023.10125697.
8. Bagaev, S., & Medvedeva, E.: Segmentation of satellite images of the earth's surface using neural network technologies. In: 2021 28th Conference of Open Innovations Association (FRUCT), pp. 15–21 (2021).
9. Kattenborn, T., Eichel, J., & Fassnacht, F.: Convolutional neural networks enable efficient, accurate and fine-grained segmentation of plant species and communities from high-resolution UAV imagery. Scientific Reports, vol. 9 (2019).
10. Tiurin, A., Vorobiev, M., Lisov, O., Andrianov, A., & Yanakova, E.: An effective algorithm for analysis and processing of satellite images for semantic segmentation. In: 2020 IEEE Conference of Russian Young Researchers in Electrical and Electronic Engineering (EIConRus), pp. 2018–2022 (2020).
11. Gonzales, C., & Sakla, W.: Semantic segmentation of clouds in satellite imagery using deep pre-trained U-nets. In: 2019 IEEE Applied Imagery Pattern Recognition Workshop (AIPR), pp. 1–7 (2019).

12. Liu, Z., Feng, R., Wang, L., Zhong, Y., & Cao, L.: D-resunet: Resunet and dilated convolution for high resolution satellite imagery road extraction. In: IGARSS 2019 - 2019 IEEE International Geoscience and Remote Sensing Symposium, pp. 3927–3930 (2019).
13. Chen, L., Bentley, P., Mori, K., Misawa, K., Fujiwara, M., & Rueckert, D.: Drinet for medical image segmentation. IEEE Transactions on Medical Imaging, vol. 37, pp. 2453–2462 (2018).
14. Grosgeorge, D., Arbelot, M., Goupilleau, A., Ceillier, T., & Allioux, R.: Concurrent segmentation and object detection CNNs for aircraft detection and identification in satellite images. In: IGARSS 2020 - 2020 IEEE International Geoscience and Remote Sensing Symposium, pp. 276–279 (2020).
15. Sedov, A., Khryashchev, V., Larionov, R., & Ostrovskaya, A.: Loss function selection in a problem of satellite image segmentation using convolutional neural network. In: 2019 Systems of Signal Synchronization, Generating and Processing in Telecommunications (SYNCHROINFO), pp. 1–4 (2019).
16. Gonzalez, J., Bhowmick, D., Beltrán, C., Sankaran, K., & Bengio, Y.: Applying knowledge transfer for water body segmentation in Peru. ArXiv, vol. abs/1912.00957 (2019).
17. Ji, S., Wei, S., & Lu, M.: Fully convolutional networks for multisource building extraction from an open aerial and satellite imagery dataset. IEEE Transactions on Geoscience and Remote Sensing, vol. 57, pp. 574–586 (2019).
18. Ibtehaz, N., Rahman, M.S.: MultiResUNet: Rethinking the U-Net architecture for multimodal biomedical image segmentation. Neural Networks 121, 74–87 (2020). https://doi.org/10.1016/j.neunet.2019.08.025
19. Yin, X.-X., Sun, L., Fu, Y., Lu, R., Zhang, Y.: U-Net-Based Medical Image Segmentation. Journal of Healthcare Engineering, vol. 2022, Article ID 4189781, 16 pages (2022). https://doi.org/10.1155/2022/4189781
20. Weng, Y., Zhou, T., Li, Y., Qiu, X.: NAS-Unet: Neural Architecture Search for Medical Image Segmentation. IEEE Access 7, 44247–44257 (2019). https://doi.org/10.1109/ACCESS.2019.2908991
21. Wang, X.-Y., Wang, T., Bu, J.: Color image segmentation using pixel-wise support vector machine classification. Pattern Recognition 44(4), 777–787 (2011). https://doi.org/10.1016/j.patcog.2010.08.008
22. Kumar, A., Sarkar, S., Pradhan, C.: Malaria Disease Detection Using CNN Technique with SGD, RMSprop and ADAM Optimizers. In: Dash, S., Acharya, B., Mittal, M., Abraham, A., Kelemen, A. (eds.) Deep Learning Techniques for Biomedical and Health Informatics, Studies in Big Data, vol. 68, pp. 99–110. Springer, Cham (2020). https://doi.org/10.1007/978-3-030-33966-1_11
23. Hasan, S.M.K., Linte, C.A.: U-NetPlus: A Modified Encoder-Decoder U-Net Architecture for Semantic and Instance Segmentation of Surgical Instruments from Laparoscopic Images. 2019 41st Annual International Conference of the IEEE Engineering in Medicine and Biology Society (EMBC), Berlin, Germany, pp. 7205–7211 (2019). https://doi.org/10.1109/EMBC.2019.8856791
24. Cheng, J., Tian, S., Yu, L., Liu, S., Wang, C., Ren, Y., Lu, H., Zhu, M.: DDU-Net: A dual dense U-structure network for medical image segmentation. Applied Soft Computing 126, 109297 (2022). https://doi.org/10.1016/j.asoc.2022.109297